\documentclass{article} 
\usepackage{iclr2026_malgai,times}

\usepackage{amsmath,amsfonts,bm}

\def\eqref#1{equation~\ref{#1}}

\def\1{\bm{1}}

\DeclareMathAlphabet{\mathsfit}{\encodingdefault}{\sfdefault}{m}{sl}
\SetMathAlphabet{\mathsfit}{bold}{\encodingdefault}{\sfdefault}{bx}{n}

\usepackage{hyperref}
\usepackage{url}

\usepackage{graphicx}
\usepackage{newfloat}
\usepackage{placeins}
\DeclareFloatingEnvironment[
          	fileext=los,
          	name=Exhibit,
]{exhibit}
\usepackage[table,dvipsnames]{xcolor}
\usepackage[normalem]{ulem}

\title{Beyond Reasoning: RL-Policy Guided LLM Inference for Efficient Strategy in Liar’s Poker\thanks{We acknowledge significant contributions by Jefferey Rosenbluth and William Wong in the early stages of this research. We also benefited from helpful discussions with Victor Haghani, Marc Lanctot, Bart de Vylder, Zachary Lipton, Noam Brown, David Buch, Adrian Weller, and Utkarsh Patange. We would also like to thank the human players who spent hours playing with the AI agent and providing us with helpful feedback. Their names and bios are listed in the appendix.}}

\author{Richard~Dewey \& János~Botyánszki\thanks{Corresponding authors} \\
Allometry Labs\\
San Francisco, CA, 94133, USA \\
\texttt{\{rich,janos\}@allometrylabs.com} \\
\And
Ciamac~C.~Moallemi \\
Graduate School of Business \\
Columbia University \\
New York, NY, USA \\
\texttt{ciamac@gsb.columbia.edu} \\
\AND
Andrew~T.~Zheng \\
Sauder School of Business \\
University of British Columbia \\
Vancouver, Canada \\
\texttt{andrew.zheng@sauder.ubc.ca}
}

\iclrfinalcopy 
\begin{document}

\maketitle

\begin{abstract}
AI researchers have long focused on poker-like games as a testbed for environments characterized by multi-player dynamics, imperfect information, and reasoning under uncertainty. While recent breakthroughs have matched elite human play at no-limit Texas hold'em, the multi-player dynamics are subdued; most hands quickly converge with only two players engaged through multiple rounds of bidding. In this paper, we present Solly, the first AI agent to achieve elite human play in reduced-format Liar’s Poker, a game characterized by extensive multi-player engagement. We trained Solly using self-play with a model-free, actor-critic, deep reinforcement learning (RL) algorithm. Solly played at an elite human level as measured by win rate (over 50\% of hands) and equity (money won) in heads-up and multi-player Liar’s Poker. Solly also outperformed frontier large language models (LLMs) on the same metrics. In one extension, when Solly's policies were provided to LLMs as a domain-specific guide for inference, LLM performance improved and token costs decreased. This result suggests that low-cost RL agents can be used to boost LLM efficiency and reasoning capabilities.
\end{abstract}

\section{Introduction}

Games often require logic, reasoning under uncertainty, and probabilistic thinking, making them an ideal environment for developing broader intelligence. Multi-player games with imperfect information form a sub-genre that has proven particularly challenging for AI modeling efforts. 

The first AI vs. human game to capture widespread attention was the 1997 chess match between IBM's Deep Blue \citep{campbell2002deep} and Garry Kasparov. A similar milestone was achieved when AlphaGo \citep{silver2016mastering}, an AI agent developed by Google DeepMind, beat Lee Sedol at the ancient Chinese board game Go in 2016. 

\begin{figure}[t]
    \centering
    \includegraphics[width=0.7\textwidth]{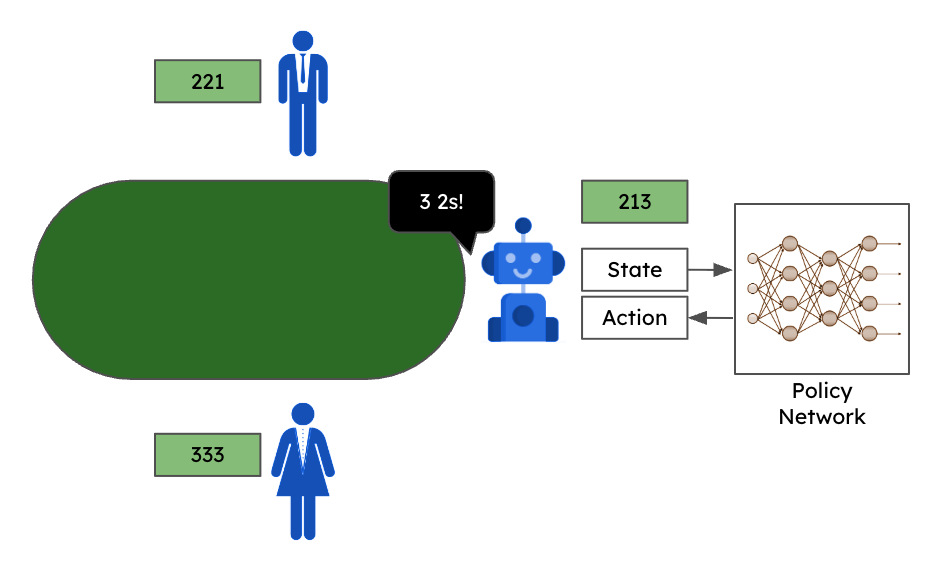}
    \caption{Reduced-format 3x3 Liar's Poker is played by bidding on the cumulative digits across all players. Solly calculates the bidding policy using a neural network and selects a move from the distribution output by it. Solly was trained via self-play.}
    \label{fig:lp_table}
\end{figure}

Both Chess and Go represent games with perfect information and no randomness. AI has also been successful in more complex strategy games such as Stratego \citep{Stratego, sokota2025stratego} that have imperfect information. There has also been significant progress building AI systems to play the flagship poker game no-limit Texas hold’em (NLTH) with DeepStack \citep{moravcik2017deepstack} and Libratus \citep{brown2018libratus} in the 2-player setting and Pluribus \citep{brown2019pluribus} in the multi-player setting.  Successful algorithms for no-limit Texas hold'em such as DeepStack \citep{moravcik2017deepstack} combined RL, search, and CFR. Pluribus \citep{brown2019pluribus}, the first AI to defeat humans at multi-player NLTH made use of RL and search, while the more general ReBel framework \citep{brown2020rebel} for 2-player zero-sum imperfect information games combines RL, search, and neural networks. More recently, \citet{studentofgames2023} introduced the Student of Games framework that uses search, self-play, and game-theoretic reasoning to learn a whole class of perfect information (e.g. Chess, Go) and imperfect information (e.g. heads-up NLTH, Scotland Yard) games.

Although AI agents have been shown to play human level in multi-player NLTH, a large percentage of NLTH hands degenerate into 2-player scenarios after the opening round of betting, as most players fold. It is, therefore, unclear how well these AI programs perform in a multi-player setting where all players participate in each betting round.

Liar's Poker, in contrast, requires multi-player engagement throughout the entire game. Often played with three or more players, the game exemplifies the challenges of extending game-solving techniques beyond two-player zero-sum settings. Liar's Poker is a popular game among financial market participants. Combining statistical reasoning with decision making under uncertainty, the game is believed to reinforce important skills for bidding in settings with imperfect information, such as auctions for spectrum rights, electricity, and government bonds. It is also thought to highlight decision-making biases like those described in \citet{tversky1974judgment} that would be costly for trading desks. We briefly introduce the rules for playing the game below and provide a formal description in Appendix \ref{appx:lp_rules}.

\section{Liar's Poker}

Figure \ref{fig:lp_table} shows a visual representation of a Liar's Poker game. In each round of Liar's Poker, players are dealt a private hand of digits. The objective is to make bids for the total number of one of the digits across all player hands. A player might bid ``four 2s'' if they believe a total count of the digit 2 among all players is at least four. Subsequent players can either challenge this bid or submit a higher bid, defined as higher in digit (e.g. ``four 3s'') or count (e.g. ``five 1s''). When all players have challenged a bid, the bidder has the option to either end the round or to submit a rebid. Any bid may be rebid once, and the round ends if all other players challenge the rebid. Once a round ends, digits are counted and the bidder receives 1 unit of reward from each player if their bid was correct or pays out 1 unit to each player otherwise.

The rebid mechanism is thought to be unique among poker-style games. Having the option to rebid is particularly valuable when a player has a strong hand. For example, a player bidding on 2s can probe to see if other players have 2s or are strong in different digit. In the event the bid on 2s is challenged, the player can then increase the bid to another digit, which often has a destabilizing effect on other players by signaling that the first bid might have been a bluff. Alternatively, if a player has a weak hand and is quickly challenged by other players, the rebid can give them an option to try another digit that opponents might feel less comfortable challenging and thereby escape a losing bid.

We will use the notation ``$H$x$D\ L$-player'' to denote a configuration of Liar's Poker with $H$ digits per player, $D$ digit cardinality, and $L$ players. For example, 8x10 2-player refers to a heads-up game in which each player's hand contains 8 digits that can range from 0-9, while a 3x3 3-player refers to a game with 3 players whose 3 digits are randomly pulled from the set $\{$1, 2, 3$\}$. A formal description of the game is provided in Appendix \ref{appx:lp_rules}. 

Liar's Poker is similar in sprit to Dudo, a dice game believed to have originated in the Inca Empire during the 15th century; similar variants of the game are Call My Bluff, Bluff, and Liar’s Dice. Dudo was solved in the 2-player setting with a classical tabular approach by \citet{OptimalDudoPlayFixedStrategy}. Methods such as counterfactual regret minimization \citep[CFR;][]{zinkevich2007cfr} and Monte Carlo CFR \citep[MCCFR;][]{lanctot2009mccfr} were the standard algorithms for poker and poker-like games until neural fictitious self-play \citep[NFSP;][]{heinrich2016nfsp} combined reinforcement learning (RL) and neural networks to approximate the state space in games with larger state spaces. \citet{gendre2019dudo} explored playing multi-player Dudo with CFR. In that work, they played a single round and evaluated against a generic baseline rather than human players. The work on Diplomacy \citep{diplomacy2022} is probably the most closely related this research. Diplomacy is a multi-player, imperfect information game with randomness and is both cooperative and competitive, so it serves as another ideal test bed for this class of games.

\section{Objectives}

In this paper, we introduce Solly, the first AI to defeat elite human players at reduced-format Liar’s Poker. Given the complexity of AI evaluation in multi-player games, we use performance against humans to evaluate the performance of Solly, as was done for Cicero \citep{diplomacy2022} and Pluribus \citep{brown2019pluribus}. Michael Lewis’ seminal book \textit{Liar's Poker} \citep{lewis1989liars} describes the high-stakes competitive games that were played on Wall Street in the 1980s; we tested Solly against some of the best players of this era. Our human participants had decades of experience, often played for thousands of dollars, and developed theory around bidding dynamics. We provide their biographies in Appendix \ref{appx:players}. Solly also played reduced-format Liar's Poker against exploiting best response RL agents and advanced LLMs. 

To the best of our knowledge, this is the first published research on training AI agents to play Liar's Poker. Our algorithmic method for Liar's Poker is based on DeepNash \citep{Stratego}, which uses a model-free RL approach with a neural network to approximate the best policy response and value function. The RL component of DeepNash is the regularized Nash dynamics (R-NaD) algorithm, which succeeded in approximating a Nash equilibrium for the 2-player game Stratego. We implement their architecture with some modifications for Liar's Poker in the multiple-player setting.

While training an agent to play multi-player Liar's Poker at the elite human level was the primary objective of this work, a second-order research goal was to develop Solly with limited compute. To that end, we play modified, smaller Liar's Poker game sizes. Specifically, we trained neural networks to play 3x3 2- and 3-player scenarios, as well as 5x5 2-player, and tested them against multiple opponent types.

\section{Methodology}

To train agents to play Liar's Poker, we used the regularized Nash dynamics (R-NaD) actor-critic algorithm of \citet{rnad-perolat21a}, which implements follow the regularized leader (FoReL) dynamics with an additional regularization term on the game reward to train agents to play sequential, imperfect information games, providing strong guarantees on convergence to Nash equilibrium in the 2-player setting. \citet{Stratego} combined R-NaD with a deep neural network architecture to build the DeepNash AI to play Stratego, a 2-player capture the flag board game. 

As shown in \citet{etessami2007complexity} and \citet{daskalakis2009computing}, computing a Nash equilibrium for three or more players is difficult and, in any case, does not provide the same worst-case guarantees as in the 2-player zero-sum setting \citep{ShohamLeytonBrown09}. While R-NaD does not guarantee convergence in a multi-player setting \citep{rnad-perolat21a}, the basic approach of regularized policy optimization via self-play generalizes naturally, and we adapt the two-player R-NaD implementation from OpenSpiel \citep{lanctot2019openspiel}, to this end. To our knowledge, this is the first study of R-NaD's performance in a multi-player setting.

To handle the $\sim$$10^{17}$ state space of our three-player Liar's Poker variant, we chose R-NaD over CFR for its memory and compute efficiency. Tabular CFR algorithms—and even sampling variants like ES-MCCFR—require exhaustive game-tree traversals that scale poorly in deep bluffing games. For instance, prior research applying CFR to Dudo \citep{neller2012} demonstrated that the time cost per iteration increases by almost two orders of magnitude for each added die. In that study, computing a single CFR training iteration for just one simulated pair of die rolls with 7 dice (a 2-vs.-5 configuration) took approximately 4.5 days. Because CFR relies on full recursive traversals, it becomes computationally infeasible for games like Liar's Poker that feature exponentially growing paths to nodes of greater depth. R-NaD circumvents this exponential tabular footprint by utilizing a neural network to smoothly generalize across the state space.

We experiment with two multi-player generalizations of R-NaD: (1) a direct extension that preserves the regularization scaling and value network architecture of 2-player R-NaD and simply collects trajectories via self-play between $N$ players, and (2) an extension which scales the opponents' regularization terms by $-\frac{1}{N-1}$ to force each entropy window's regularized game to remain strictly zero-sum, and expands the value network to output one value per player position. Empirically, both methods produced the same long-term best response score, and results reported in the body correspond to the direct approach.

Our network consists of a simple, fully-connected MLP, which takes as input a fixed-size representation of the game state and outputs a vector of logits specifying a distribution over possible actions: either a feasible bid or a challenge. The MLP consists of a shared torso of two hidden layers with separate policy and value heads. We provide further details around the training configurations and compare the two multi-player approaches in Appendix~\ref{appx:training}.

\section{Results}

Finding a Nash equilibrium is difficult in the multi-player setting and might not produce a winning strategy, given that other players may not be playing the equilibrium. Therefore, our primary form of evaluation was game-play against elite human players, which has not previously been attempted for this class of games. We also compute a best response score, which gives a method for comparison across game sizes and architectures. Finally, we evaluate the AI against LLMs by playing against both general-purpose and reasoning models.

\subsection{Elite Human Evaluation}

We evaluated Solly against elite human players following the approach used by researchers in no-limit Texas hold'em \citep{brown2019pluribus}. Solly's opponents played Liar's Poker on Wall Street during the 1980s and 1990s, had extensive experience playing for high stakes, and have thought deeply about the game and its bidding dynamics. We include their names and backgrounds in Appendix \ref{appx:players}. 

We compare both the average win rates and player equity or total dollar winnings, a common performance metric in the poker literature.

We selected a group of seven players and played games in different configurations, both online and in-person, over the course of three months. We first tested our agent heads-up against humans and found that the agent was comparable to the best Liar's Poker players. We report the complete results of the various games in Exhibit \ref{ex:results_summary}.

\begin{exhibit}[t]
	\begin{center}
		{\small
        \renewcommand{\arraystretch}{1.3}
			\begin{tabular}{ccccc}
				\textit{\textbf{3x3 2-Player}} \\
			 	\hline 
				  &  &  & \textbf{Solly Scaled} & \textbf{Solly Scaled}\\
 				\textbf{Opponent(s)} & \textbf{Hands Played} & \textbf{Solly Win \%} & \textbf{Equity} & \textbf{Std. Error}\\
				\hline   
				Chat-GPT 4.1 & 1,000 & 60\% & \$19 & $\pm \$3$ \\ 
				OpenAI o3 & 1,000 & 55\% & \$9 & $\pm \$3$  \\ 
                Elite Humans & 100 & 48\% & -\$4 & $\pm \$10$ \\ 
                Best Response & 1,000 & 44\% & -\$12 & $\pm \$3$ \\ 
				\hline
				 & & & \\
				 & & & \\
			\end{tabular}
            \begin{tabular}{ccccc}
				\textit{\textbf{5x5 2-Player}} \\
			 	\hline 
				  &  &  & \textbf{Solly Scaled}& \textbf{Solly Scaled}\\
 				\textbf{Opponent(s)} & \textbf{Hands Played} & \textbf{Solly Win \%} & \textbf{Equity} & \textbf{Std. Error}\\
				\hline  
				Elite Humans & 100 & 55\% & \$10 & $\pm \$10$ \\ 
				\hline
				 & & & \\
				 & & & \\
			\end{tabular}
			\begin{tabular}{ccccc}
				\textit{\textbf{3x3 3-Player}} \\
			 	\hline 
				  &  &  & \textbf{Solly Scaled}& \textbf{Solly Scaled}\\
 				\textbf{Opponent(s)} & \textbf{Hands Played} & \textbf{Solly Win \%} & \textbf{Equity} & \textbf{Std. Error}\\
				\hline  
				2 Elite Humans & 100 & 54\% & \$17 & $\pm \$15$ \\ 
				\hline
			\end{tabular}
        }
	\end{center}
	\caption{Solly's performance against various opponents, scaled to show the expected equity (money) Solly won per 100 hands.}
    \label{ex:results_summary}
\end{exhibit}

We played 100 hands of heads-up 3x3 Liar's Poker against elite humans and Solly won 48\% of those hands. We next played 100 hands of heads-up 5x5 Liar's Poker against elite humans and Solly won 55\% of those hands. In both cases, we randomly selected five elite humans and played 20 hands against each in a mix of in-person and online games.

One of our key contributions is the performance of Solly in the multi-player setting. We played 100 hands against a subset of elite players in-person. In the 3-player setting, we used the 3x3 configuration featuring two humans against the AI agent. Our null hypothesis is that Solly's mean reward is zero. Said differently, on average we expect Solly to be no different than elite humans players. In our 3x3 3-player game, Solly won against humans with an average score of 0.17 and a standard error of 0.15. Based on these results, we cannot reject our null hypothesis with a two-tailed T-test (N=$100$, T=$1.10$, P=$0.27$). 

Compared to humans, Solly tended to use the rebid feature to bluff more than humans. In the 3x3 3-player setting, Solly rebid in roughly 33\% of hands, while humans each only used the rebid in about 8\% of hands. This is atypical of what human players have long considered an optimal strategy. We discuss performance of players broken out by hand quality and win type in Appendix \ref{appx:hand_quality}. 

\subsection{Best Response Agent} 

A standard measure of agent performance is the strength of a best response policy against the AI. A best response policy for a player or AI is one that maximizes that player's return against all other players \citep{ShohamLeytonBrown09}, in this case maximizing the reward against one or more copies of the Solly agent. This method provides an estimate of Solly's exploitability, and the best response score we report here is the average reward per round for the best response agent. 

Best response training differs from the main training methodology in a few key ways. First, the best response agent is trained specifically to exploit a fixed training checkpoint of Solly, whereas Solly is trained to improve itself using self-play against its current, evolving policy state at each step. Since computing a best response against a fixed policy effectively reduces the complex, multi-agent environment to a stationary Markov Decision Process (MDP), a compact network architecture is sufficient. Second, each best response agent is trained in a fixed position of the game (initial bidder, second bidder, etc.), while Solly's training in R-NaD takes data from all player positions into account when updating the network. Since we are using a neural network to calculate the best response, it is an approximation and should be viewed as a lower bound; it is possible that an optimal exploiter could find better exploits.

To approximate exploitability, we trained Deep Q-Network (DQN) best response agents using the OpenSpiel framework.  Each DQN agent was trained for 1 million environment steps with a 100,000-transition replay buffer from which batches of 32 transitions were sampled for gradient updates. The network architecture consisted of an MLP with three hidden layers of 64 units each,\footnote{We also tested a higher-capacity DQN agent (four times the hidden layer size and a batch size of 128) and found it did not yield a higher best response score, confirming that our initial architecture was not limited by network capacity.} optimized via Stochastic Gradient Descent (SGD) with a Mean Squared Error (MSE) loss function and a learning rate of 0.1; because Liar's Poker features bounded, low-value rewards, this comparatively high learning rate is needed to prevent the optimization from stalling. We used an epsilon-greedy exploration schedule (decaying to 0.1) and explicitly disabled exploration ($\epsilon = 0.0$) during evaluation to ensure a deterministic best response. To investigate whether the position of the player affects the strength of the best response policy, we trained a separate agent for each position. In the 3-player case, one DQN agent learned while two instances of Solly played with a fixed policy. We evaluated every 5,000 environment steps with 1,000 rounds of play for each best response agent position, and report on the rolling average results of the last 10 evaluation points.

\begin{figure}[t!]
\includegraphics[width=.49\textwidth]{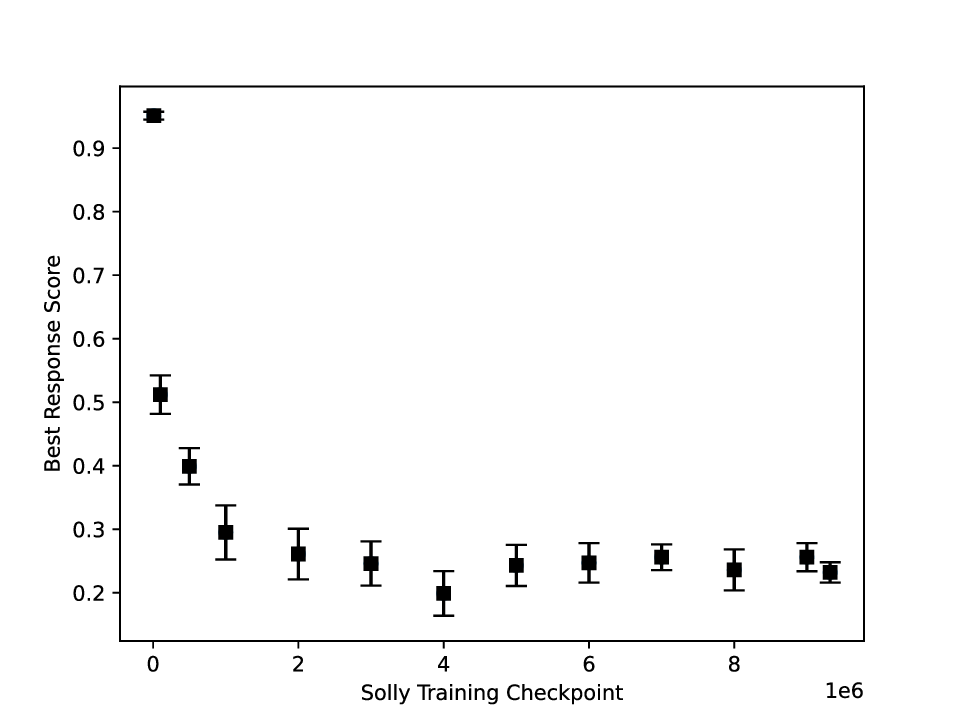}
\includegraphics[width=.49\textwidth]{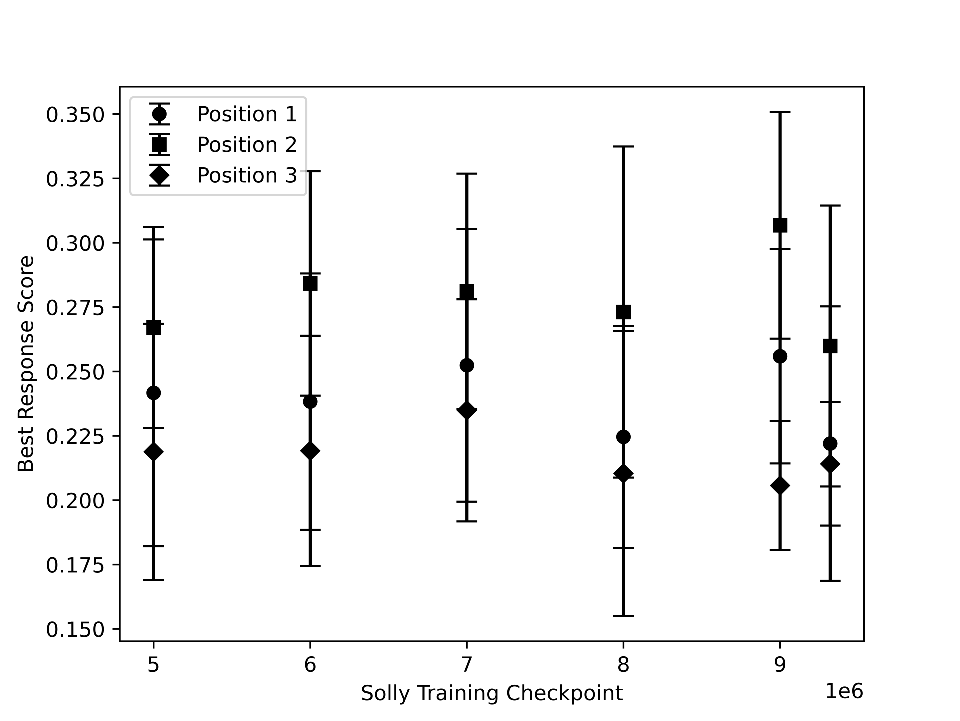}
\caption{Best response scores for the 3x3 3-player configuration. A lower score means a better quality Solly agent. The first panel shows the average best response score for agents trained to play against various Solly training checkpoints across all player positions. The Solly agents improve (become less exploitable) as training progresses. The second panel shows the scores of the exploiting agents playing in each of the three player positions, zoomed in on checkpoints 5M and above.}
\label{fig:BR_3x3x3}
\end{figure} 

We show the results for the 3x3 3-player game in Figure \ref{fig:BR_3x3x3}. The first panel shows that best response agents perform well against a Solly trained for only 10k steps (average reward $\sim$1.0). However, for more advanced training checkpoints of Solly, the best response agent's reward drops to $\sim$0.25 units. For context, a player who wins every round through a successful bid (earning 2 units each time) would have a best response score of 2. The decreasing trend indicates that the best response agents had an increasingly hard time exploiting Solly as Solly gained more experience through self-play. The second panel of the figure depicts the best response score separately for each player position at checkpoints 5M and above. We find no material differences between the scores at the three player positions, suggesting that Solly is equally difficult to exploit regardless of the player order.

\subsection{Large Language Models - large sample game-play vs OpenAI models}
\label{sec:openai}
The restricted size of the 3x3 game is a rich testbed for evaluating both trained agents and large language models (LLMs) on Liar's Poker due to the reduced complexity of the bid sequences. We conducted preliminary testing of LLMs playing Liar's Poker using a stateful session.

In our first prompt, we submitted an HTML-structured version of the Salomon Brothers rules. We explicitly prompted the LLM to play the reduced 3x3 game using the rebidding feature but no other extensions. We provided a randomly-generated hand to the LLM agent at the beginning of each round and a reminder of the correct response options. At the end of each round, we announced the total digit count, winner, and win type (successful bid or successful challenge) to the LLM before starting the next round. We limited the games to 100-round batches to refresh the context, providing the LLM with the instructions at the beginning of each batch. 

We played Solly against Chat-GPT 4.1 for 1,000 hands, and Solly won 60\% of rounds. 83\% of Solly's wins were achieved through successful bidding, while only 34\% of the LLM wins were achieved through a successful bid (see Exhibit \ref{ex:hand_quality} in the Appendix). The LLM never used the rebid option.

Solly also played 1,000 hands of Liar's Poker against the OpenAI o3 reasoning model and won 55\% of hands. Similarly to what we observed against GPT-4.1, Solly won 70\% of those hands through successful bidding. However, the reasoning model performed better in bidding compared to its non-reasoning counterpart; it won 53\% of its games through successful bidding.

Both models used a deterministic bidding strategy based on calculations of probabilities for the unknown digits, and they generally assume no bluffing. However, GPT-4.1's reasoning around probability is vague, and it does not explicitly report calculating the binomial probabilities needed to guide bidding. The o3 model also uses expected values to choose a bid. These factors likely explain the performance differences between the o3 and GPT-4.1 models.

We conducted several preliminary performance tests in the multi-player setting with Solly and LLMs in the spirit of exploration and completeness. One primary test of interest was to play Solly against two instances of the o3 reasoning model. We found the second o3 model to have an advantage, winning equity of 15 points per 100 rounds, with Solly losing slightly (-3 points per 100 rounds). The second o3 model was at a disadvantage, losing -12 points per 100 rounds.

There is likely a subtle form of collusion between the LLMs because they use the same strategy, which is to play deterministically according to maximizing expected value, conditional on their hand, while taking each player's bid as highly representative of their hand with no consideration of bluffing. The LLM preceding Solly has an advantage because the other LLM's bid correlates highly with that agent's hand, therefore ``leaking'' information about its hand to its counterpart. It would be akin to two humans committing to playing a strategy in advance and never deviating. This demonstrates a known challenge of multi-player games. Multiple equilibria may exist, and they are not interchangeable; unless all players play the same equilibrium policy, their combined policy is likely not an equilibrium, leading to sub-optimal payoffs.

Our intuition is that humans who are able to adapt would easily defeat these LLMs in a multi-player setting, as they would quickly observe that the LLM is leaking information to them. The pre-trained Solly is not able to adapt in real-time or use test-time compute (TTC) to gain insights from previous hands. Although the LLM is generating a reasoning output and using TTC, we confirmed by inspecting reasoning traces that it is not adapting to the players, thinking about the best action with respect to future hands (i.e. bluffing now might be rewarded later), or tracking game history to identify systematically exploitable behavior. The LLMs play deterministically, and their inability to adapt or make use of the game history makes them suboptimal agents in our view.

A final aspect to note is that the LLMs appeared to play very conservatively. We can only speculate on the drivers of this without full access to the model weights and training corpus. The LLM could be influenced by the corpus of standard poker literature online, which often advocates folding most hands and only playing aggressively with top hands. The LLMs might have taken a more deterministic approach and only bid when the probabilities of winning exceeded 50\%, failing to randomize their play and bluff, which Solly learned through extensive self-play. Providing the LLMs with more curated, language-specific feedback during game play and using LLMs with greater inference-time compute are interesting directions for future research.

\begin{figure}[t!]
\centering
\includegraphics[width=.95\textwidth]{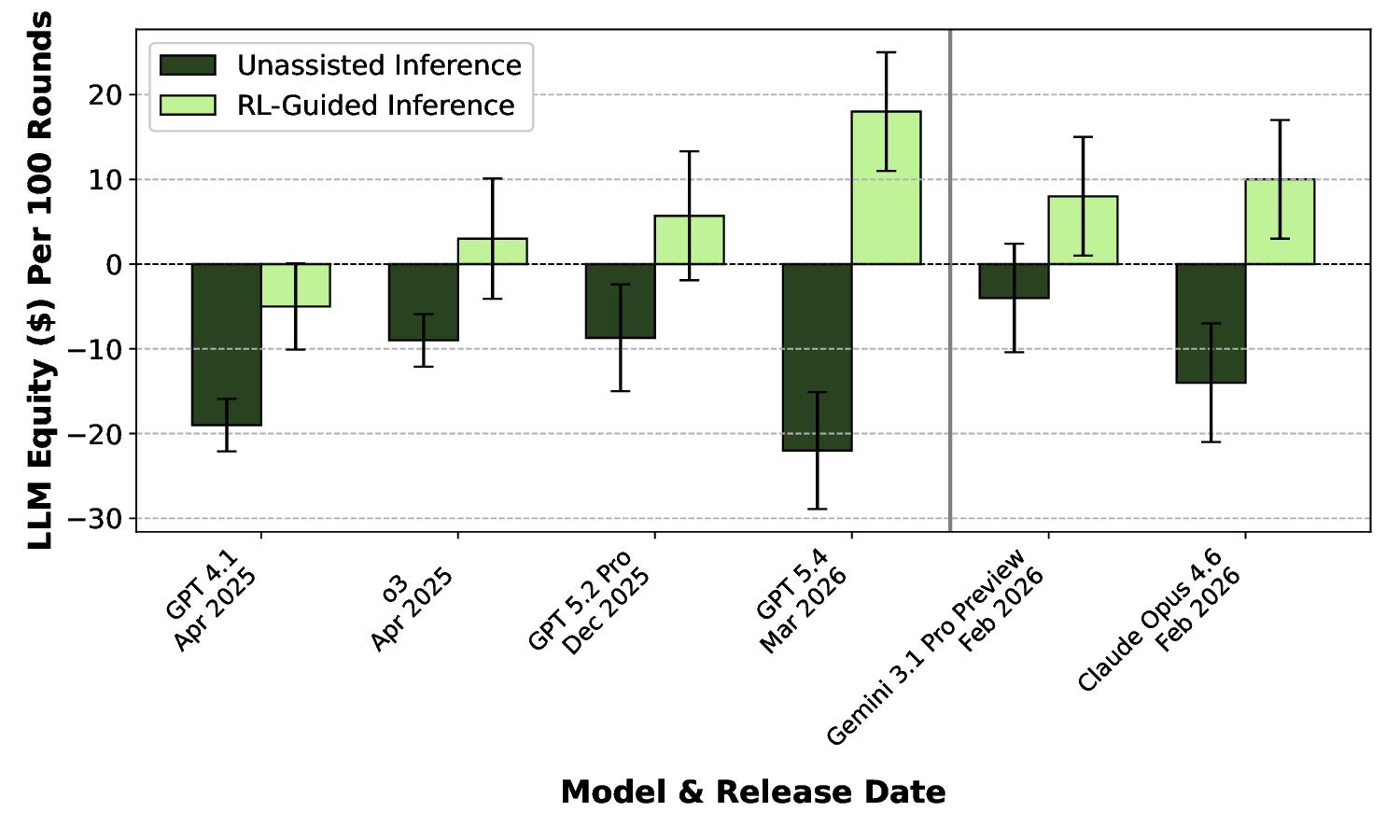}
\caption{Equity won by various frontier LLMs when playing against Solly in a 3x3 2-player setting. The left side shows OpenAI models sorted by release date, and we provide the leading model from Google and Anthropic in the right side for comparison. RL-policy guided inference differs from unassisted in that Solly's policies are added to the LLM prompt at each decision point. Unassisted results from Section \ref{sec:openai} are included for GPT 4.1 and o3. While all models underperform Solly in the unassisted mode, these models consistently improve in equity performance when provided with RL-policy guidance, with newer models performing the best in the assisted mode.}
\label{fig:LLM_equity}
\end{figure}

\subsection{RL-Guided Inference: low-cost RL-learned policies as input for frontier LLMs}

LLMs generally do not have sufficient internal knowledge of games to perform complex strategic reasoning, and they benefit from domain-specific policy input \citep{hwang2025largelanguagemodelsdevelop}. Systems that combine humans with AI could prove to be superior to humans or AI alone, as was shown in Chess \citep{feng2023chessgpt}, Poker \citep{huang2024pokergpt, maugin2025spingpt} and Go \citep{wang2025empowering}.

To investigate the benefit of RL agent and LLM cooperation, we exposed Solly's policy to various frontier LLMs in a 3x3 2-player game; see Appendix \ref{appx:llm_integration} for complete implementation details. We played both a standard, unassisted version of Liar's Poker and one in which Solly's stochastic policy was shared with the language model at each decision point; we will refer to this latter approach as RL-guided LLM inference.

LLM equity outcomes are displayed in Figure \ref{fig:LLM_equity}. The overall trend is that pro-tier, high-reasoning models perform better on standard (unassisted) games. More interestingly, however, the RL-policy guided inference approach improved LLM performance for most of the frontier LLMs.  

Qualitatively, we found that frontier LLMs used the R-NaD policy at inference-time as a strong guide to their decision-making, reducing the amount of reasoning tokens used and improving performance. They considered Solly's recommended actions and layered on further reasoning before deciding on their next move. Their ability to outplay Solly (gain positive equity) demonstrates the incremental strategic benefit of the language models' reasoning abilities; blind compliance with the RL policy would be expected to result in an even matchup. This result underscores the advantage of combining the reasoning capabilities of language models with domain-specific learned intelligence.

Furthermore, providing in-context policy guidance had tangible cost benefits. Table \ref{tab:reasoning_tokens} shows the average reasoning tokens generated by each model in our two scenarios. We show that, across all frontier models tested in this work, the hybrid inference approach reduces thinking token counts. We see the largest differences in the most recent high-performance models (Opus 4.6 and GPT 5.4). We speculate that their superior routing ability \textemdash{} stopping reasoning when provided with clear guidance \textemdash{} compared with older models may explain the higher difference between the assisted and non-assisted thinking token usage.  

\begin{table}[]
    \centering
    \renewcommand{\arraystretch}{1.3}
    \small
    \begin{tabular}{cccc}
        & \multicolumn{2}{c}{\textbf{Thinking Token Usage (tokens / request})} & \\
         Model & In-Context Policy Guidance & Unassisted & Difference \\
         \hline
         o3 & 1066 & 1383 & \textcolor{ForestGreen}{+30\%} \\ 
         GPT-5.2 Pro & 617 & 643 & \textcolor{ForestGreen}{+4\%} \\ 
         GPT-5.4 & 932 & 4049 & \textcolor{ForestGreen}{+334\%} \\ 
         Gemini 2.5 Pro & 163 & 221 & \textcolor{ForestGreen}{+36\%} \\ 
         Gemini 3.1 Pro Preview & 144 & 127 & \textcolor{Mahogany}{-12\%} \\ 
         Claude Sonnet 4.6* & 297 & 325 & \textcolor{ForestGreen}{+9\%} \\ 
         Claude Opus 4.6* & 92 & 280 & \textcolor{ForestGreen}{+203\%} \\ 
        \hline
        \multicolumn{4}{l}{*Anthropic's API does not break out reasoning tokens. We display total output tokens instead.} \\
    \end{tabular}
    \caption{Thinking token usage is higher in the unassisted scenario across all frontier models included in our study with the exception of Gemini 3.1 Pro Preview, indicating strong cost savings when LLMs are provided RL-policy guidance. Each request represents one LLM decision point.}
    \label{tab:reasoning_tokens}
\end{table}

\section{Discussion}

In this paper, we demonstrated human-level performance by an AI agent at reduced-format Liar's Poker in the heads-up and multi-player settings. This is the first human-level performance in a multi-player poker game that requires full engagement of all players in every betting round. It also includes a rebid feature, not found in other games and conducive to bluffing, which is thought to favor humans. We developed a process for training agents in an n-player setting and achieve elite human-level performance. Our training was done with relatively small neural networks and limited computing resources, making it accessible to researchers outside of the major AI labs.

Recently, there has been increasing interest in using LLMs to play games. We explored the use of LLMs to play reduced-format Liar's Poker and discussed their shortcomings. One interesting line of inquiry for humans vs. machines is in training time vs test-time duration (e.g., thinking time for humans, total generation time for LLMs). We surveyed our human players and found that they played approximately 40,000 and 50,000 hands in their life.\footnote{The calculation for this was 40 hands per day, 3 time per week, 50 weeks a year for 7-9 years.} We recorded our games and found they spent an average of 10 to 30 seconds thinking before making a move. ChatGPT 4.1 provided responses in 1 to 3 seconds, while o3 provided responses in 10 to 40 seconds, and presumably neither model had prior Liar's Poker training. In contrast, Solly was trained on billions of hands and used nearly zero TTC. \citet{brown2019pluribus} find that using more TTC delivers extraordinary improvements in performance in six-player no-limit poker, and indeed, most breakthrough game AIs used some form of test-time computation \citep[see e.g.,][]{campbell2002deep, silver2016mastering, brown2018libratus, sokota2025stratego}. It is very likely that Solly would have benefited from a technique like Monte Carlo tree search (MCTS) at test-time, but we leave that to future research efforts to explore. 

While preliminary, our results in providing LLMs with RL-policy guidance support the need for further research in using/training LLMs with domain-specific rewards/policies. Our low-cost RL agent increased both performance and token efficiency across most frontier LLMs, suggesting that such an agent would be beneficial in other domains even when constrained to a certain model, provider, or inference budget.

In scaling an agent to play a larger configuration of Liar's Poker, we suspect that there are efficiency gains from using the smaller games to bootstrap learning in larger games. We also expect optimizations around reward scaling, representation abstraction, test-time search, behavioral cloning, and truncating the look-back period might offer further improvements in performance. We discuss scaling further in Appendix \ref{appx:scaling}.

\subsubsection*{Acknowledgments}
The authors gratefully acknowledge the computational resources and support provided by the Briger Family Digital Finance Lab at the Columbia Business School.

We acknowledge significant contributions by Jefferey Rosenbluth and William Wong in the early stages of this
research. We also benefited from helpful discussions with Victor Haghani, Marc Lanctot, Bart de Vylder, Zachary
Lipton, Noam Brown, David Buch, Adrian Weller, and Utkarsh Patange. We would also like to thank the human
players who spent hours playing with the AI agent and providing us with helpful feedback. Their names and bios
are listed in Appendix \ref{appx:players}.

\bibliography{iclr2026_malgai}
\bibliographystyle{iclr2026_conference}

\appendix
\section{Appendix}

\subsection{Player Backgrounds}
\label{appx:players}

\subsubsection*{Jeffrey Rosenbluth}
Jeffrey Rosenbluth is a founding member of Math for America's Board. He is a private investor and former Managing Director and Head of Fixed Income Arbitrage at Salomon Brothers, Inc. Dr. Rosenbluth serves as an overseer of the School of Engineering and Applied Science at the University of Pennsylvania. He is a trustee of Harvey Mudd College. He graduated from the University of Pennsylvania with a BSE in Computer Science, earned a Ph.D. in Mathematics from the Courant Institute at NYU, and holds an MBA from the University of Chicago.

\subsubsection*{Victor Haghani}
Victor Haghani started his career in 1984 at Salomon Brothers in bond research. He moved to the trading floor in 1986 and, shortly after, became a managing director in the bond arbitrage group run by John Meriwether. In 1993, Victor was a co-founding partner of Long-Term Capital Management (LTCM). He established and co-ran its London office. He is the founder and current CIO of Elm Wealth. 

\subsubsection*{Pete Muller}
Pete Muller is an American investor, singer-songwriter, and philanthropist. He began his career at BARRA before moving on to Morgan Stanley, where he founded the quantitative trading group PDT, which was later spun out into PDT Partners, a stand-alone hedge fund. In his spare time, Pete is an accomplished musician, recording seven studio albums and touring around the world. He also designs crossword puzzles and is an avid game player. He has won Poker Night on Wall Street three times and has made it to the final table on the World Poker Tour and in a World Series of Poker event.

\subsubsection*{Eric Rosenfeld}
Eric Rosenfeld is a retired hedge fund manager. He recently taught an advanced Fixed Income course at Yale School of Management and MIT Sloan. Rosenfeld was also a co-founder of LTCM. His previous role was at Salomon Brothers, where he was appointed head of the Government Trading Department and a member of the Salomon Brothers Executive Committee. From 1979 to 1984, Rosenfeld was an assistant professor of Finance at Harvard Business School.

\subsubsection*{Aaron Brown}
Aaron Brown is a long-time poker player, financial trader, risk manager, and author. In the 1980s, he led a group of quants (before that was a word) in applying systematic Liar's Poker strategies on trading floors to gain respect and cash. He is the author of ``The Poker Face of Wall Street,'' ``A World of Chance'' (with Reuven and Gabrielle Brenner), ``Red-Blooded Risk,'' ``Financial Risk Management for Dummies,'' and the forthcoming ``Wrong Number.'' He holds degrees in Applied Mathematics from Harvard and Finance and Statistics from the University of Chicago. He writes regular columns for Bloomberg and Wilmott Magazine and hosts the ReasonTV Wrong Number video series.

\subsubsection*{Larry Bernstein}
Larry Bernstein is currently the managing member of Amber Mountain, a private investment firm specializing in fixed-income trading. He was a trader at Salomon Brothers from 1989 to 1998. Larry is the host of the What Happens Next Podcast. He graduated from the University of Pennsylvania's Wharton School of Business. 

\subsection{Official Salomon Brothers Liar's Poker Rules}
\label{appx:lp_rules}

For a full description of Liar's Poker rules, visit \url{https://elmwealth.com/wp-content/uploads/2021/11/liars.poker-rules.pdf}. Here are some excerpts to illustrate the game rules we implemented (note that our implementation supports any digit cardinality and hand length, not just 8x10, and we only used the rebid extension):

\textit{To begin the game, each player obtains a random eight digit number. The most common method is to have each player choose a bill (of US currency, generally a one dollar bill). A player's number for that round is the serial number on the bill selected.}

\textit{Play begins as one player makes the opening bid. A typical bid might be ``5 sevens.'' This means that the player estimates that the total number of sevens in all players' numbers is at least 5. He need not have all 5 sevens in his own number. The turn then passes clockwise to the next player on the left. For his turn, each player must either make a stronger bid or challenge the previous bid. A bid is stronger if it calls for at least the same number of occurrences of a higher rank (e.g., ``5 nines'') or a greater number of occurrences (e.g., ``6 threes''). The zero is considered the highest rank (usually referred to as ``ten'' as in ``7 tens'').}

\textit{Eventually, a bid will be challenged by all the remaining players. At this time, each player reveals how many of the selected rank he has. If the total number equals or exceeds the number bid, the bidder wins one unit from each of the other players. If the bid is not made, then the bidder loses one unit to each of the other players. Whether or not the bid is made, the final bidder is the first bidder in the next round.}

\textit{The most important enhancement of the basic game is the right of rebidding. If a player's bid is challenged by all of the other players, he has the option of playing this bid or of making a new, stronger bid. However, only one rebid is allowed.}

\textit{If this new bid is also challenged by all the players, the bidding then stops and this new bid is the final bid. If this new bid is not challenged by all of the remaining bidders and one of them makes a stronger bid, the bidding continues with each player---including the bidder who just made a rebid---having the right to rebid if challenged all the way around.}

\textit{The allowance of rebids greatly extends the strategic scope of Liar's Poker. The necessity to bluff and determine others' bluffs is a major feature of the game.}

A formal representation of this game follows.

A round of Liar's Poker is played between \( L \) players, indexed by \( \ell \in \{1, \ldots, L\} \). At the beginning of the round, each player receives a private hand, or ``SLIP,'' of $H$ digits, each taking a value chosen uniformly at random from $\{ 1 \ldots D \}$. More compactly, each player's hand can be represented by a random vector of consisting of counts for each digit \( X_{\ell} = (X_{\ell 1}, \ldots, X_{\ell,D}) \), where $X_\ell$ drawn from a multinomial distribution with $H$ trials, $D$ categories, and probability $1/D$ of drawing any given category in a given trial. The original game is played using the serial numbers on dollar bills (i.e., $D = 10$ and $H=8$, where a 0 digit on the bill means 10), but the game soon changed to randomly generated numbers.

Aggregating across all players gives the global count vector \( S = \sum_{\ell=1}^L X_\ell \), where \( S_j \) denotes the total number of occurrences of digit \( j \) across all hands. A bid is defined as an ordered pair \( (q, r) \), interpreted as the claim that \( S_r \ge q \); for example, the bid $(4, 7)$ (“four sevens”) corresponds to the claim that the digit seven appears at least four times in the union of all players' hands. At the beginning of the round, one player will propose an opening bid. Players then take turns choosing an action, which may be one of two options: either to ``challenge'' the current bid, signaling that they do not believe the bid's corresponding inequality holds; or to issue a stronger bid \( (q', r') \succ (q, r) \), where we define the ordering \( (q', r') \succ (q, r) \) iff \( q' > q \) or \( (q' = q \text{ and } r' > r) \).

Eventually, some bid $(q, r)$ will be challenged by all players. At this point, the player who proposed the challenged bid may choose either to proceed with a count, in which all players reveal their hands, resolving the round and computing payout depending on whether the event $S_r \geq q$ holds; or they may choose to rebid, i.e., to propose a stronger bid $(q', r') \succ (q, r)$, and continue playing. The rebid option is only available if the challenged bid $(q, r)$ did not itself result from a rebid. Like other bids, a rebid must be challenged by all other players for the round to end. Upon resolution, players are paid out based on the final bid $(q, r)$. If the bid is correct, i.e., \( S_r \ge q \), the bidder gains one unit from each opponent; otherwise, the bidder pays one unit to each opponent.

\subsection{Neural Network Architecture and Configurations}
\label{appx:training}

The information tensor of the Liar's Poker game, which serves as the policy network's input layer, encodes the actor's hand and position, a one-hot encoded history of each player's bids, and a one-hot encoded history of each player's challenges, a bit representing the rebid state, and a bit representing whether or not the trajectory is terminal. The output layer is given by the number of maximum number of allowed bids plus one for the challenge move. As the game size determines the number of possible bids and challenges, these tensors scale by game size and set a minimum reasonable network architecture configuration. 

We used a multi-layer perceptron (MLP) consisting of a shared torso with 2 hidden layers of size 256 and separate policy and value heads. For context, the state representation tensor of the 3x3 3-player game is of length 170. We set the maximum trajectory length to 10 for 2-player scenarios and 15 for 3-player scenarios. We set all action probabilities below 3\% to zero during game play and discretized the non-zero values to a 32 value grid (see the supplementary materials to \citet{Stratego}) to avoid low-probability, catastrophic moves allowed by the softmax function during inference. We used the Adam optimizer without momentum ($\beta_1 = 0$) to prevent cyclical dynamics and divergence \citep{balduzzi18a, gidel19a}. We tested and found that a learning rate of 0.00005 was ideal for the game sizes we trained. 

To expand R-NaD to a multi-player game, we implemented both a direct and a generalized extension of the algorithm. In the direct baseline, we kept the value head a scalar and maintained the identical regularization scaling from the 2-player version. In the generalized extension, we expanded the value head to dimension $N$ to track independent player values, and scaled the opponents' regularization terms by $-\frac{1}{N-1}$ to force each entropy window's regularized game to remain strictly zero-sum. Empirically, both methods produced the same long-term best response score, though the generalized zero-sum implementation converged faster. This acceleration occurs because the $N$-dimensional head provides independent baselines that reduce policy gradient variance, while the scaled penalty preserves the game's zero-sum saddle point, avoiding inefficient, potentially cyclical learning dynamics. Regardless, as R-NaD's outer loop updates the fixed policy target at the end of each window, the KL divergence penalty naturally decreases, and the zero-sum reward structure of the original game (either {2, -1, -1} or {-2, 1, 1} depending on the outcome) ultimately dominates the total reward. Our finding that the direct extension of R-NaD does indeed converge toward an approximate equilibrium as target policies are updated is an interesting, if unexpected, result of this work.

\subsection{Hand Quality} 
\label{appx:hand_quality}

Another measure of agent quality is performance conditional on different canonical hands. A three-of-a-kind hand (for example [2, 2, 2]) is considered the strongest hand, while a mixed hand of one digit each (for example [2, 3, 1]) is considered the weakest. 

Exhibit \ref{ex:hand_quality} shows the performance of humans and Solly with each hand type. While the baseline trend is that players tend to win more on stronger hands, one interesting result is that humans in the 3-player setting underperformed with 2-of-a-kind hands, while Solly did less so. As the majority of hands in the 3x3 configuration are 2-of-a-kind, this particular win rate has a disproportional influence on the final equity results. 

Win type is another insightful dimension. In 2-player games, winning by either bid or challenge results in a uniform +1 reward for the round, whereas the reward becomes asymmetric in multi-player games (+2 for a win by bid, +1 for a win by challenge). Solly's equity edge over elite humans in the multi-player setting is partially attributable to its higher win rate from correct bids, for which there was an out-sized reward. 

\begin{exhibit}[!t]
	\begin{center}
		{\small
			\begin{tabular}{cccc|cc|c}
				\textit{\textbf{3x3 2-Player}} \\
			      \hline 
                &&&&\textbf{Win By} &\textbf{Win By} &\textbf{Overall}\\
 				\textbf{Player} & \textbf{1 Digit} & \textbf{2 Digits} & \textbf{3 Digits} &
                \textbf{Bid} & \textbf{Challenge} & \textbf{Win Rate}\\
				\hline   
				Elite Humans & 40\% & 49\% & 92\% & 62\% & 38\% & 52\%  \\
                Solly & 34\% & 47\% & 100\% & 42\% & 58\% & 48\% \\ 
				\hline
				\rowcolor{lightgray} Chat-GPT 4.1 & 25\% & 40\% & 72\% & 34\% & 66\% & 40\% \\ 
                \rowcolor{lightgray} Solly & 36\% & 61\% & 91\% & 83\% & 17\% & 60\% \\ 
                \hline
				OpenAI o3 & 22\% & 48\% & 75\% & 53\% & 47\% & 45\% \\ 
                Solly & 10\% & 62\% & 90\% & 70\% & 30\% & 55\% \\ 
                \hline 
                \rowcolor{lightgray} Best Response & 34\% & 56\% & 95\% & 53\% & 47\% & 55\% \\ 
                \rowcolor{lightgray} Solly & 29\% & 45\% & 70\% & 35\% & 65\% & 45\% \\ 
                \hline
                \\
                \\

			\end{tabular}
            \begin{tabular}{cccc|cc|c}
				\textit{\textbf{3x3 3-Player}} \\
			 	\hline 
                &&&&\textbf{Win By} &\textbf{Win By} & \textbf{Overall}\\
 				\textbf{Player(s)} & \textbf{1 Digit} & \textbf{2 Digits} & \textbf{3 Digits} &
                \textbf{Bid} & \textbf{Challenge} & \textbf{Win Rate}\\
				\hline    
				Elite Humans & 50\% & 36\% & 78\% & 31\% & 69\% & 46\% \\ 
                Solly & 62\% & 48\% & 71\% & 50\% & 50\% & 54\%\\ 
				\hline
			\end{tabular}
        }
	\end{center}
	\caption{Solly performance broken down by hand quality and the percent of wins due to a bid or challenge. 1 Digit refers to a mixed hand with one of each digit, 2 Digits is one with 2-of-a-kind, and 3 Digits is a 3-of-a-kind hand. The majority of hands in the 3x3 configuration are 2-of-a-kind.}
    \label{ex:hand_quality}
\end{exhibit}

\subsection{Scaling}
\label{appx:scaling}

We believe scaling the R-NaD algorithm to play the complete (8x10) Liar's Poker agent is straightforward with enhancements to the network architecture, tuning of hyperparameters, state representation abstraction, and more compute. As the number of players, digits, and/or hand length increase, the state space grows, and network training efficiency becomes more important. 

One inherent challenge of scaling to a larger game size is that the strength of the reward signal is increasingly diluted as the length of rounds increases. Unreasonably high bids are often rewarded during self-play because a (sub-optimal) opponent bids even higher and loses; on the other hand, a reasonable bid that wins is rewarded with the same amount as a challenge (or just slightly more with 3+ players). Reward scaling for training agents to play the complete game could address this issue.

\begin{figure}[t!]
\includegraphics[width=.49\textwidth]{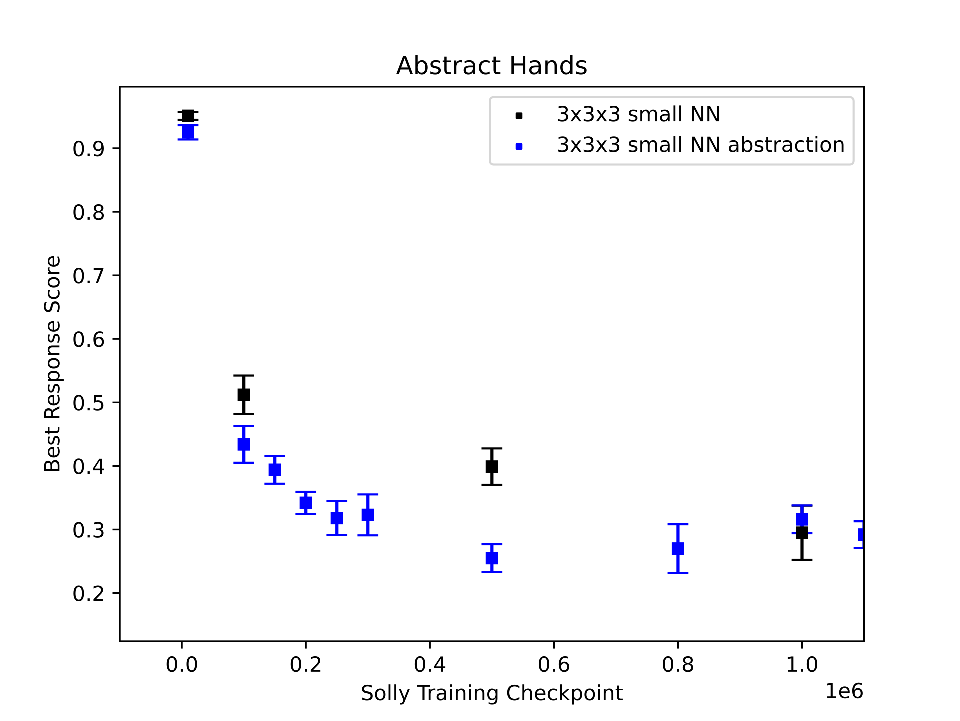}
\includegraphics[width=.49\textwidth]{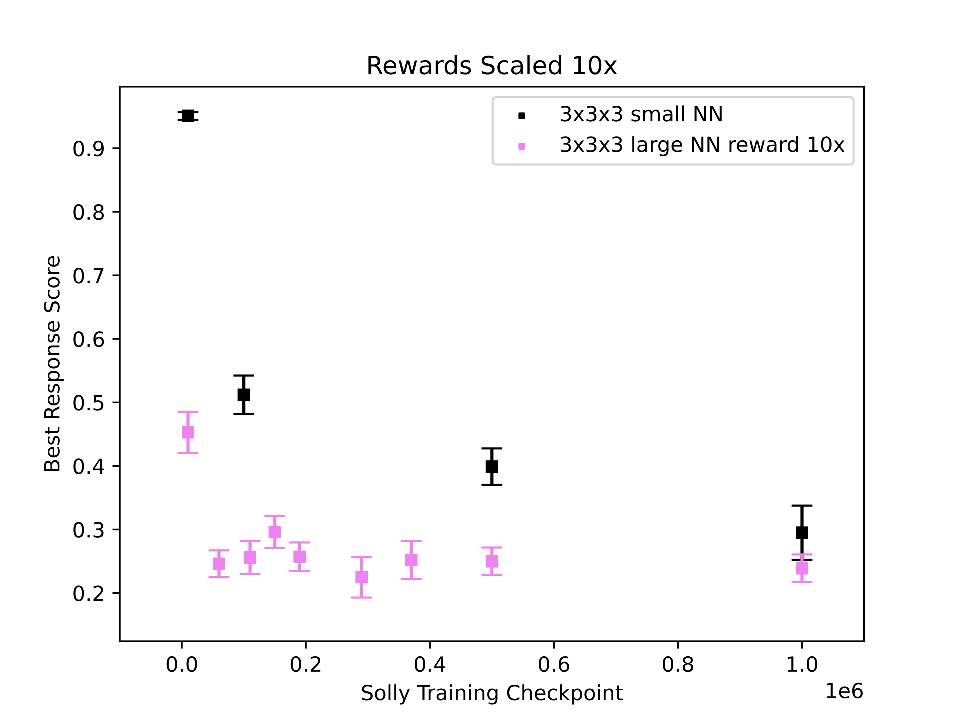}
\caption{Best response scores for 3x3 3-player demonstrating two proposed scaling techniques. In the first panel, we rewrite the Liar's Poker environment to encode hands as digit counts, training on abstract (``canonical'') hands rather than explicit digits. We compare this agent to the original 3x3 3-player agent used for play against elite humans. In the second panel, we compare against an agent trained with a deeper MLP (7 layers of 512 neurons each) and rewards scaled by a factor of 10.}
\label{fig:BR_3x3x3_scaling}
\end{figure}

In an effort to determine the potential for scaling our approach to the full 8x10 game, we explored the use of reward scaling during training, hand abstractions (i.e. grouping strategically identical hands together, which is common in the poker literature), deeper MLPs commensurate with the size (if not sophistication) of models used in the game playing literature, and tuning several hyperparameters. Figure \ref{fig:BR_3x3x3_scaling} shows the improvements for two such configurations and gives us comfort that with straightforward architecture improvement and additional resources, it is possible to scale our algorithm to play the complete Liar's Poker game with similar performance results.

\subsection{LLM Integrations}
\label{appx:llm_integration}

To integrate Solly's policy with LLMs, we built application code with the ability to switch between OpenAI, Google, and Anthropic APIs. The core Liar's Poker functionality and LLM prompts were identical between all model types and both unassisted and policy-guided modes. In the initial request, we provided the rules of the game and instructions specifying game scenario (3x3 2-player), bidding order, and expected response format. For RL-guided inference, we also required the LLM to consider the policy provided at each step. Where available, we provided a seed and kept temperature at the default value of 1.0 to allow for creative reasoning. We played 200 rounds of Liar's Poker for all new scenarios, and we include the 1,000 hands of unassisted play from Section \ref{sec:openai} in our reporting. 

For OpenAI models, we used the openai Python SDK and the ``responses'' API. This allowed for a stateful interaction with the service; we provided the previous interaction ID to maintain the state of the session. Temperature and seed are not available in this SDK. We used a `high' reasoning effort setting.

For Google Gemini models, we first integrated them using the Model Context Protocol \citep[MCP;][]{mcp_specification_2024} to probe an agentic approach. However, we found that a significant amount of thinking budget was routed to a Flash model (even when specifying the Pro model) and that Pro thinking tokens were suppressed; we suspect the model used Flash for orchestrating the MCP calls and spent significant thinking tokens doing so. Ultimately, we used the google-genai package with REST requests to ensure that thinking tokens were used primarily for the game and results with other frontier LLMs are comparable. We provided the previous interaction ID to maintain the state of the session. We set the seed to 104729, thinking level to `high,' and temperature to 1.0.

For Anthropic models, we used the anthropic Python package. The seed parameter is not available in this library. We set the temperature to the default value of 1.0, thinking mode to `adaptive,' routing to explicit, and maximum output tokens to 16,000. We provided a full history of interactions with each request in order to maintain the state of the session.  

While the LLMs did not have visibility into Solly's private hands, there is the possibility that, by providing Solly's state-specific policies, the LLM could, in principle, infer how Solly might play in similar states. Inspecting the reasoning summaries, we did not see any evidence for this. For one, the LLM was not told it was playing against the agent or an opponent with the same policy as the agent. Nor is there a high chance that Solly would end up in the same exact state for which the LLM had previously seen a policy; and without knowing Solly's private hand, the model would have to make assumptions about its opponent's hand in order to use that historical information. We therefore believe that the risk of leaking information to boost performance is minimal in this scenario. 

\end{document}